\documentclass[]{fairmeta}
\usepackage{wrapfig}
\usepackage{tabularx}
\usepackage{textcomp}
\usepackage{stfloats}
\usepackage{url}
\usepackage{verbatim}
\usepackage{graphicx}
\usepackage{titlesec}
\usepackage{tocloft}
\usepackage{adjustbox}
\usepackage{multirow}
\usepackage{pifont}
\usepackage{tikz}
\usepackage{comment}
\usepackage{amsmath,amssymb} 
\usepackage{colortbl}  
\usepackage[dvipsnames]{xcolor}         
\usepackage{booktabs} 
\usepackage{hyperref}
\usepackage{graphicx}    
\usepackage{subcaption} 
\usepackage{booktabs}
\usepackage{amsmath} 
\usepackage{multirow} 
\usepackage{booktabs} 
\usepackage{subcaption} 
\usepackage{todonotes}
\usepackage{natbib}
\usepackage{makecell}
\usepackage{tcolorbox,enumitem,ragged2e}
\usepackage{enumitem}

\tcbset{
  promptbox/.style={
    colback=black!2, colframe=black!20, boxrule=0.3pt,
    arc=1mm, left=6pt, right=6pt, top=6pt, bottom=6pt}
}

\RequirePackage{xspace}
\makeatletter
\DeclareRobustCommand\onedot{\futurelet\@let@token\@onedot}
\def\@onedot{\ifx\@let@token.\else.\null\fi\xspace}

\makeatother

\definecolor{adptorange}{RGB}{248, 205, 172}
\definecolor{cmpblue}{RGB}{189, 215, 238}
\definecolor{cmpblue}{RGB}{189, 215, 238}

\definecolor{our_red}{RGB}{232,157,160}
\definecolor{our_blue}{RGB}{136,206,230}
\definecolor{our_orange}{RGB}{246,200,168}
\definecolor{our_green}{RGB}{178,211,164}

\definecolor{attn_code0}{RGB}{247,215,200}
\definecolor{attn_code1}{RGB}{238,169,139}
\definecolor{mlp_code0}{RGB}{204,201,221}
\definecolor{mlp_code1}{RGB}{102,95,153}

\definecolor{dark_green}{rgb}{0, 0.5, 0}
\definecolor{dark_red}{rgb}{0.8, 0.2, 0.2}
\definecolor{soft_red}{rgb}{1.0, 0.4, 0.4}
\definecolor{light_blue}{rgb}{0.2, 0.5, 1.0}

\usepackage{algpseudocode}
\usepackage{algorithm}

\definecolor{token_blue}{RGB}{84, 120, 140}

\usepackage{pifont}       
\usepackage{bbding}       
\usepackage{fontawesome}
\usepackage{xspace}

\usepackage{float}
\usepackage{siunitx}        
\usepackage{microtype}      
\usepackage{algorithm}     
\usepackage{algorithmicx}  
\usepackage{algpseudocode} 
\usepackage{cleveref}      
\usepackage{enumitem}
\usepackage[inkscapelatex=false]{svg}
\usepackage{animate}
\usepackage{etoc}          

\definecolor{darkgreen}{rgb}{0.15, 0.75, 0.15}
\definecolor{cvprblue}{rgb}{0.21,0.49,0.74}
\definecolor{lightblue}{rgb}{0.90, 0.95, 0.99}

\algrenewcommand\algorithmicrequire{\textbf{Input:}}
\algrenewcommand\algorithmicensure{\textbf{Output:}}

\newcommand{\systemname}{\textsc{Inferix}\xspace}

\title{
  Inferix: A Block-Diffusion based Next-Generation Inference Engine for World Simulation
}
\author{Inferix Team}

\abstract{
    World models serve as core simulators for fields such as agentic AI, embodied AI, and gaming, capable of generating long, physically realistic, and interactive high-quality videos. Moreover, scaling these models could unlock emergent capabilities in visual perception, understanding, and reasoning, paving the way for a new paradigm that moves beyond current LLM-centric vision foundation models. A key breakthrough empowering them is the semi-autoregressive (block-diffusion) decoding paradigm, which merges the strengths of diffusion and autoregressive methods by generating video tokens in blocks-applying diffusion within each block while conditioning on previous ones, resulting in more coherent and stable video sequences. Crucially, it overcomes limitations of standard video diffusion by reintroducing LLM-style KV Cache management, enabling efficient, variable-length, and high-quality generation. 
    
    Therefore, Inferix is specifically designed as a next-generation inference engine to enable immersive world synthesis through optimized semi-autoregressive decoding processes. This dedicated focus on world simulation distinctly sets it apart from systems engineered for high-concurrency scenarios (like vLLM or SGLang) and from classic video diffusion models (such as xDiTs). Inferix further enhances its offering with interactive video streaming and profiling, enabling real-time interaction and realistic simulation to accurately model world dynamics. Additionally, it supports efficient benchmarking through seamless integration of InterVBench, a new fine-grained evaluation benchmark tailored for minute-long video generation scenarios. We hope the community will work together to advance Inferix and foster world model exploration.

}

\date{\today} 
\metadata[Code]{\url{https://github.com/alibaba-damo-academy/Inferix}}

\begin{document}
\thispagestyle{firstheader}
\maketitle

\section{Introduction}

\begin{figure}
    \centering
    \includegraphics[width=0.9\linewidth]{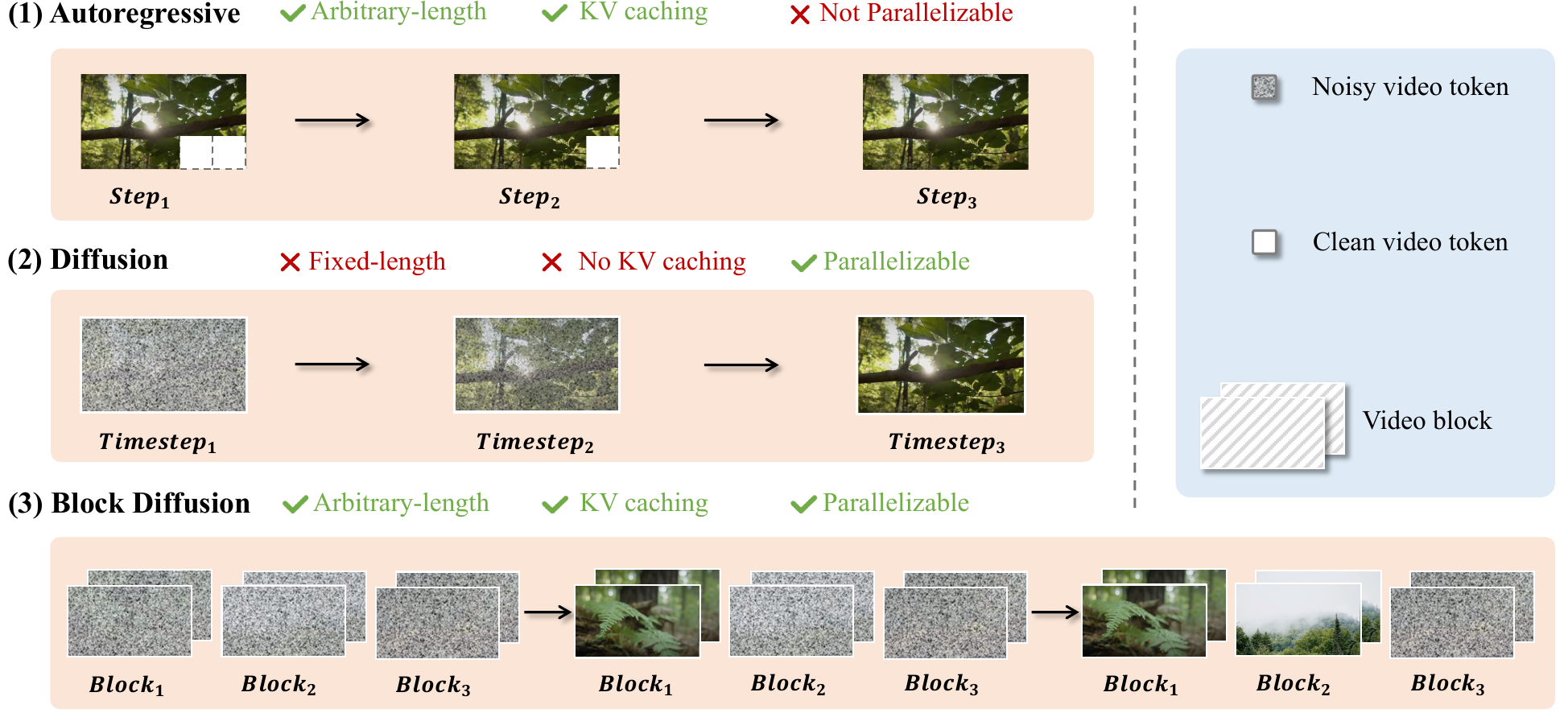}
    \caption{\textbf{Architecture comparison.} AR vs. Diffusion vs. Block
Diffusion (Semi-AR). Block Diffusion combines the strengths of both AR and Diffusion, enabling arbitrary-length generation, KV caching, and high parallelizability within each block.}
    \label{fig:arch_comparison}
\end{figure}

World models are capable of generating interactive, long-form, and physically plausible video sequences. Most current video diffusion models \cite{wan2025wan} rely on the Diffusion Transformer (DiT) \cite{peebles2023scalable}—which uses bidirectional attention without KV caching. While this enables parallelized generation and controllability, decoding is inefficient and restricted to fixed lengths. In contrast, AR-based frameworks \cite{wang2024loong} support variable-length generation and KV Cache management, but their generation quality lags behind video diffusion, and decoding is not parallelizable. Importantly, block diffusion \cite{huang2025self,teng2025magi} interpolates between AR and diffusion by reintroducing LLM-style KV Cache management, enabling efficient, variable-length, and high-quality generation, as shown in Figure~\ref{fig:arch_comparison}. 

\begin{figure}
    \centering
    \includegraphics[width=0.9\linewidth]{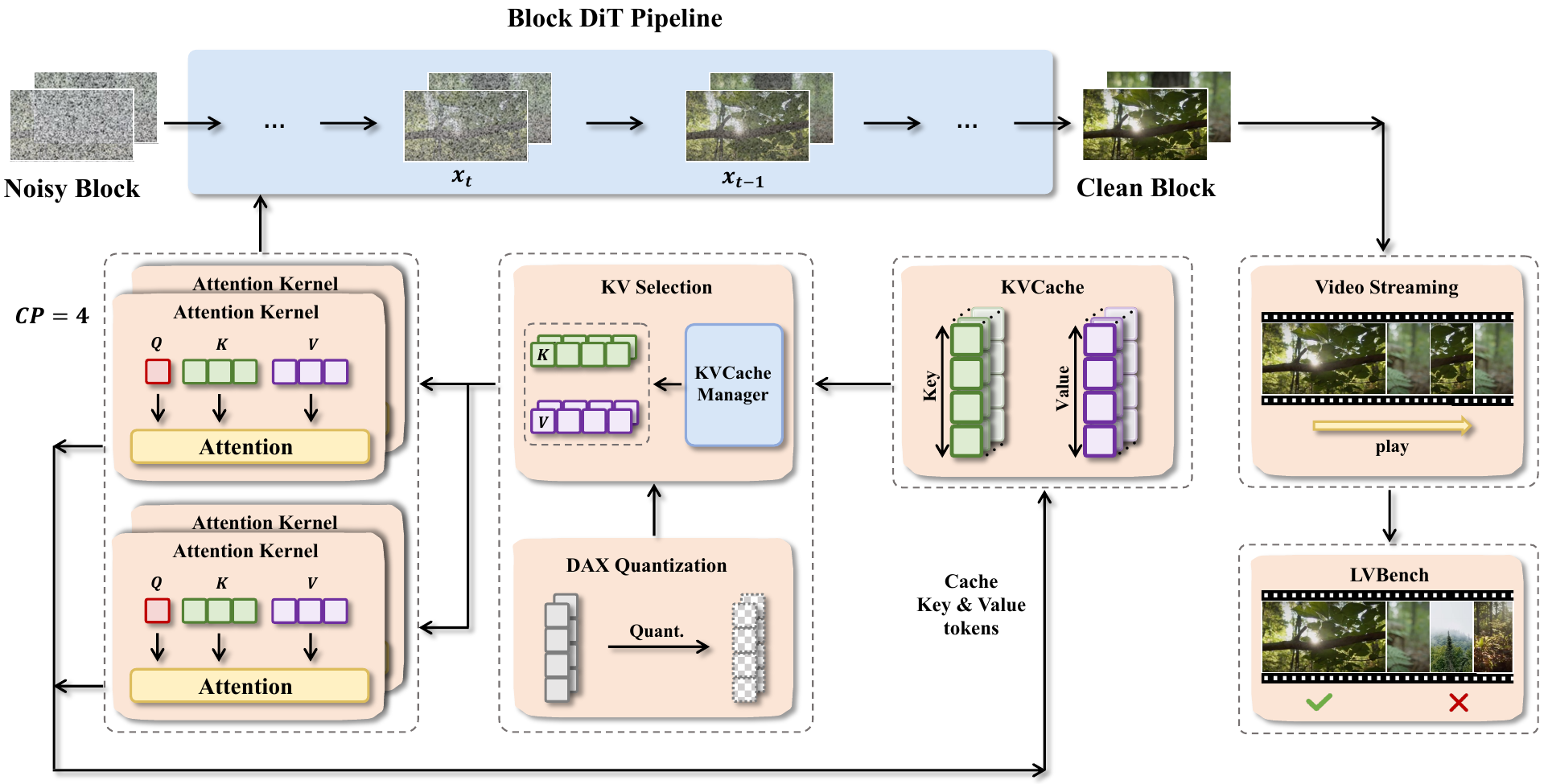}
    \caption{\textbf{Framework of Inferix.} To enhance the efficiency of block diffusion models, INFERIX provides a set of interconnected components: efficient parallel strategies, block-wise KV Cache management, DAX~\cite{dax2025} quantization, real-time video streaming, and fine-grained video evaluation.}
    \label{fig:framework}
\end{figure}

The overall framework of Inferix is illustrated in Figure \ref{fig:framework}. The model generates a clean video block from noise via iterative denoising. Crucially, the attention mechanism at each step leverages a global KV Cache containing context from previously generated blocks. After a new block is generated, its KV information is used to update the cache, providing context for subsequent blocks. This generate-and-cache loop facilitates efficient, arbitrary-length video generation.

A new paradigm inevitably brings forth new infrastructure and fundamental research, just as the LLM era gave rise to vLLM~\cite{kwon2023efficient} \& SGLang~\cite{zheng2024sglangefficientexecutionstructured}, the Visual Diffusion era to xDiT~\cite{fang2024xdit} and FastVideo~\cite{fastvideo2024}, and the Post-training era to OpenRLHF \cite{hu2025openrlhfeasytousescalablehighperformance} and verl~\cite{Sheng_2025}, etc. Now, the world model era also demands its own dedicated inference engine, and Inferix is purpose-built as a next-gen inference engine, empowering immersive world synthesis via optimized semi-autoregressive decoding paradigm.

Key features of Inferix are as follows:
\begin{itemize}[leftmargin=*]
    \item \textbf{Next-Generation Inference Paradigm}: A block diffusion framework built for immersive world synthesis at scale.

    \item \textbf{Efficient Long Video Generation Benchmarking}: Integrated with InterVBench, a fine-grained benchmark for minute-long videos with dedicated metrics to evaluate long-range coherence.

    \item \textbf{Video Streaming}: Basic video streaming capabilities for generated content, with both RTMP and WebRTC supported as streaming protocols.

    \item \textbf{Continuous Prompt Support}: Enable dynamic narrative control with different prompts for different video segments.

    \item \textbf{Advanced KV Cache Management}: Intelligent memory management for persistent world simulation.

    \item \textbf{Distributed World Synthesis}: Support multiple parallelism for large-scale immersive environment generation.

    \item \textbf{Built-in Profiling}: Performance monitoring and analysis capabilities with enhanced diffusion model profiling.

\end{itemize}

\section{Challenges in Inference of World Simulation}

During the inference of world simulation, the world models need to generate long-form video sequences. Moreover, for current world models and video generation models, their model size is pretty large. The large model size and long-form video sequences bring unprecedented pressure to storage and computing.

\textbf{For storage}, the usage of KV Caches is the main bottleneck. In world simulation, the KV Caches of former blocks need to be stored as the context for the generation of current and future blocks, which is important to ease the drifting and forgetting problem~\cite{zhang2025framecontextpackingdrift} when generating long video sequences. However, these KV Caches will consume a large amount of GPU memory. Therefore, how to make KV Cache management efficient is important for the inference of world simulation. Some advanced techniques that have been studied in LLM inference need to be brought to the inference of world simulation, such as PageAttention~\cite{kwon2023efficient}, offload~\cite{sheng2023flexgenhighthroughputgenerativeinference, lee2024infinigenefficientgenerativeinference}, KV Cache compression~\cite{liu2024kivi,li2024snapkvllmknowslooking}, and so on.

\textbf{For computation}, the large model size and extremely lone video sequences increase the amount of computation greatly. For example, it will consume about 6,800 seconds when generating a 5-second video with Wan2.1 14B in a single NVIDIA H20. For the inference of world simulation, the computation will be much more heavier due to the longer context. Therefore, it's significant to accelerate the computation of world simulation to make it accessible. There are several methods that can be taken to achieve this: quantization to utilize low-bit computation~\cite{zhao2024viditq,li2024svdquant}, sparse attention~\cite{yang2025sparse, zhangspargeattention}, decreasing the denoising steps~\cite{yin2024improveddistributionmatchingdistillation, gu2025blade}, leveraging the redundancy during inference~\cite{liu2025timestepembeddingtellsits,zhang2025fastvideogenerationsliding}, utilizing distributed computation~\cite{fang2024pipefusionpatchlevelpipelineparallelism, fang2024uspunifiedsequenceparallelism}, and so on.

\section{Framework Design}

\subsection{Parallelism}
To accelerate the inference process and minimize per-GPU memory footprint, \systemname employs a suite of parallelism techniques tailored for long sequence models. These include Ulysses-style sequence parallelism~\cite{ulysses-sp}, which partitions independent attention heads across multiple GPUs to relieve memory pressure while preserving computational efficiency, and Ring Attention~\cite{liu2024ringattention, yang2025contextparallelismscalablemilliontoken}, which enables scalable attention computation over long sequences by distributing attention operations in a ring topology. Depending on the selected attention mechanism, ring attention can either pass queries or pass keys and values, leading to different performance profiles. \systemname selects the most suitable parallelism strategy based on model architecture, network topology, and communication overhead. This adaptive approach ensures optimal resource utilization and performance across deployment scenarios.

\subsection{KV Management}

Block-Diffusion-based models leverage KV caches to accelerate the generation process. To support KV cache access from various kinds of models, \systemname provides a unified KV management interface backed by block-wise KV memory management. To maintain scalability in the face of future-time models requiring both sliding-window access patterns and selective global KV context dependency, the KV management system preserves the extensibility of flexible KV fetching methods including both range-based chunked access and index-based selective fetch. Latent store used in Multi-latent Attention (MLA)~\cite{liu2024deepseek} and offloading to main memory for GPU memory optimization are also supported to keep the KV management future-proof.

\subsection{Models and Pipelines}
The \systemname framework is designed to support a variety of block diffusion models, currently MAGI-1 \cite{teng2025magi}, CausVid~\cite{yin2025causvid}, and Self Forcing \cite{huang2025self} as examples. These models differ in their foundations: CausVid and Self Forcing are built upon Wan2.1, a 5-second full-attention base diffusion video model, while MAGI-1 is trained from scratch with a distinct infrastructure. To efficiently accommodate this diversity, \systemname first abstracts their shared computational patterns into a generalized inference pipeline. Building on this abstraction, we then design and integrate several key components, such as a sophisticated KV Manager and a suite of parallel strategies, to significantly boost inference performance. Users are welcome to integrate their own models with these abstractions and interfaces.

\subsection{System Profiling}
\systemname provides a built-in performance profiling mechanism that enables end-to-end visibility into resource utilization during inference. The profiler includes three key characteristics:

\textbf{Near zero overhead.} The full profile only incurs minimal overhead of less than $5\%$, compared with no profiling.

\textbf{Highly customizable.} In addition to GPU usage and system-wide metrics, \systemname allows users to add custom metrics during inference. Users can define custom metrics via lightweight hooks or callbacks that execute inline with inference, enabling domain-specific measurements.

\textbf{Easy to use.}  The profiler exposes both Python decorator and context manager. The Python decorator enables declarative profiling of individual functions, while the context manager supports block-level instrumentation for broader code regions with almost no code change.

\subsection{Video Streaming}

When generating long videos or executing world simulation, it is important to enable dynamic narrative control with different signals for different video chunks. These signals include prompts, motions, inputs from peripherals and so on. For example, when inference with CausVid, \systemname supports generating a long video whose different video chunks are controlled by different prompts specified by users. If a different prompt is given when generating a new video chunk, \systemname will clear the cross-attention cache to eliminate the influence brought by the former prompt.

\section{InterVBench}

\begin{table}[t]
\centering
\caption{Overview of the datasets used for constructing InterVBench.}
\setlength{\tabcolsep}{10pt}
\renewcommand{\arraystretch}{1.2}
\resizebox{\linewidth}{!}{
\begin{tabular}{lcc}
\toprule
\textbf{Dataset} & \textbf{Video Number} & \textbf{Object Classes} \\
\midrule
DanceTrack   & 66  & Humans (66, 100\%) \\
GOT-10k      & 272 & Humans (177, 65\%)~ Animals (54, 20\%)~ Environment (41, 15\%) \\
HD-VILA-100M & 117 & Humans (47, 40\%)~ Animals (35, 30\%)~ Environment (35, 30\%) \\
ShareGPT4V   & 545 & Humans (381, 70\%)~ Animals (82, 15\%)~ Environment (82, 15\%) \\
\midrule
\textbf{InterVBench}   & \textbf{1000} & \textbf{Humans (671, 67\%)~ Animals (171, 17\%)~ Environment (158, 16\%)} \\
\bottomrule
\end{tabular}
}
\label{tab:datasets}
\end{table}

\subsection{Dataset}
To address the challenge of generating minute-long videos, we construct \textbf{InterVBench}, a large-scale benchmark comprising 1,000 long-form videos collected from diverse open-source sources. As summarized in Table~\ref{tab:datasets}, we select high-resolution videos exceeding 50 seconds from DanceTrack~\citep{sun2022dancetrack}, GOT-10k~\citep{huang2019got}, HD-VILA-100M~\citep{xue2022advancing}, and ShareGPT4V~\citep{chen2024sharegpt4v}. 

To ensure comprehensive temporal coverage and linguistic diversity, we employ GPT-4o as a data engine to generate detailed captions every 2–3 seconds. The prompting pipeline and examples are included in Subsection ~\ref{app:prompts}. To guarantee annotation quality, we adopt a rigorous human-in-the-loop validation framework across all stages: (1) \textit{data sourcing}, where annotators filter out low-quality or unsuitable clips; (2) \textit{chunk segmentation}, where human reviewers ensure temporal coherence and eliminate transition artifacts; and (3) \textit{caption verification}, where annotators refine automatically generated descriptions for semantic accuracy and temporal alignment. Each validation stage involves at least two independent reviewers to maintain inter-rater reliability. Finally, the curated dataset is divided into an 80/20 train–evaluation split.

\subsection{Metrics}
Evaluating long-form video generation requires assessing both spatial fidelity and temporal stability. Prior studies~\citep{li2025longdiff, lu2024freelong} introduce \textit{drift penalties} to quantify degradation over time, focusing on identity consistency~\citep{han2025show} and perceptual robustness~\citep{xie2025moca}. Inspired by the Mean Absolute Percentage Error (MAPE) and Weighted MAPE~\citep{kim2016new, de2016mean}, we propose a unified metric, \textbf{Video Drift Error (VDE)}, which measures relative quality changes across the temporal axis. 

Building upon VDE, we design five complementary metrics for long-horizon video evaluation: (1) \textit{VDE-Clarity}, assessing temporal drift in image sharpness; (2) \textit{VDE-Motion}, quantifying smoothness of motion dynamics; (3) \textit{VDE-Aesthetic}, capturing consistency of visual appeal; (4) \textit{VDE-Background}, measuring spatial stability of scene layouts; and (5) \textit{VDE-Subject}, detecting identity drift in primary subjects. Lower scores in each indicate stronger temporal consistency. Following prior benchmarks~\citep{guo2025long,cai2025mixture}, we also integrate five complementary quality dimensions from VBench~\citep{huang2024vbench}: Subject Consistency~$\uparrow$, Background Consistency~$\uparrow$, Motion Smoothness~$\uparrow$, Aesthetic Quality~$\uparrow$, and Image Quality~$\uparrow$. Together, these metrics form a comprehensive protocol for evaluating long video generation models.

\subsection{Prompts for InterVBench's Data Engine}
\label{app:prompts}

\begin{center}
\begin{tcolorbox}[promptbox, before skip=0pt, after skip=0pt]
\textbf{Role.} Act as a professional video content analyst. Describe a given video frame in English.\\
\textbf{Context.} The previous frame was described as: \emph{"\{previous\_description\}"}. Use this as context to ensure temporal coherence.\\
\textbf{Instruction.} Write a single, descriptive paragraph that: 
\begin{itemize}[itemsep=1pt,leftmargin=10pt]
\item Identifies the \textbf{main subject}, their specific actions, and expressions. 
\item Describes the \textbf{environment and background}, including setting and lighting. 
\item Highlights the \textbf{cinematic quality}, such as composition, color palette, and atmosphere (e.g., tense, serene, spectacular). 
\end{itemize}
\textbf{Constraints.} Output must be \textbf{one coherent paragraph}, written in natural language prose, without bullet points or numbered lists.\\
\textbf{Return.} The paragraph description of the current frame.
\end{tcolorbox}
\end{center}

\section{Development Roadmap}

\begin{itemize}
    \item Support more complex KV Management, with flexible block-sparse attention

    \item Support finetuning a pretrained video generation model (Diffusion to Semi-AR) \& distill models into few steps~\cite{huang2025selfforcingbridgingtraintest,yin2024improveddistributionmatchingdistillation}

    \item Support high-concurrency deployment 

    \item Support more complex distributed inference 

    \item Improve video streaming usage and performance

    \item Support more advanced real-time, interactive streaming capabilities
    
\end{itemize}

\section{Conclusion}

We develop \systemname, a block-diffusion based next-generation inference engine for world simulation, which integrates some important features and a new benchmark for long video generation. The inference engine takes the differences between block-diffusion generation and former generation paradigms as the starting point, which intends to make the researches in world model and long video generation more convenient. \systemname unifies the inference interfaces of different block-diffusion models and apply several efficient inference techniques, which are important to improve ease of use. The InterVBench integrated in \systemname aims to evaluate the quality of long video generation precisely and efficiently, which is also valuable for the development of world model.

For future works, more efficient inference techniques specific to block-diffusion generation will be taken into considerations, which includes sparse attention, feature cache, step distillation and so on. We hope \systemname will become a useful tool for this.

\beginappendix

\section{Contributions and Acknowledgments}

We are a joint team from Zhejiang University \& Hong Kong University of Science and Technology \& Alibaba DAMO Academy \& Alibaba TRE.  All current contributors of Inferix are listed in alphabetical order by their last names.

We warmly welcome everyone to join our virtual team and together harness the collective power of community.
\\

\textbf{\Large Contributors}:

\textcolor{violet}{Tianyu Feng}

\textcolor{violet}{Yizeng Han}

\textcolor{violet}{Jiahao He}

\textcolor{violet}{Yuanyu He}

\textcolor{violet}{Xi Lin}

\textcolor{violet}{Teng Liu}

\textcolor{violet}{Hanfeng Lu}

\textcolor{violet}{Jiasheng Tang}

\textcolor{violet}{Wei Wang}

\textcolor{violet}{Zhiyuan Wang}

\textcolor{violet}{Jichao Wu}

\textcolor{violet}{Mingyang Yang}

\textcolor{violet}{Yinghao Yu}

\textcolor{violet}{Zeyu Zhang}

\textcolor{violet}{Bohan Zhuang}

\clearpage
\bibliographystyle{plain}
\bibliography{paper}

@inproceedings{ulysses-sp,
author = {Jacobs, Sam Ade and Tanaka, Masahiro and Zhang, Chengming and Zhang, Minjia and Aminadabi, Reza Yazdani and Song, Shuaiwen Leon and Rajbhandari, Samyam and He, Yuxiong},
title = {System Optimizations for Enabling Training of Extreme Long Sequence Transformer Models},
year = {2024},
isbn = {9798400706684},
publisher = {Association for Computing Machinery},
address = {New York, NY, USA},
url = {https://doi-org.lib.ezproxy.hkust.edu.hk/10.1145/3662158.3662806},
doi = {10.1145/3662158.3662806},
booktitle = {Proceedings of the 43rd ACM Symposium on Principles of Distributed Computing},
pages = {121–130},
numpages = {10},
location = {Nantes, France},
series = {PODC '24}
}

@inproceedings{
liu2024ringattention,
title={RingAttention with Blockwise Transformers for Near-Infinite Context},
author={Hao Liu and Matei Zaharia and Pieter Abbeel},
booktitle={The Twelfth International Conference on Learning Representations},
year={2024},
url={https://openreview.net/forum?id=WsRHpHH4s0}
}

@misc{yang2025contextparallelismscalablemilliontoken,
      title={Context Parallelism for Scalable Million-Token Inference}, 
      author={Amy Yang and Jingyi Yang and Aya Ibrahim and Xinfeng Xie and Bangsheng Tang and Grigory Sizov and Jeremy Reizenstein and Jongsoo Park and Jianyu Huang},
      year={2025},
      eprint={2411.01783},
      archivePrefix={arXiv},
      primaryClass={cs.DC},
      url={https://arxiv.org/abs/2411.01783}, 
}

@article{xie2025moca,
  title={MoCA: Identity-Preserving Text-to-Video Generation via Mixture of Cross Attention},
  author={Xie, Qi and Ma, Yongjia and Di, Donglin and Gao, Xuehao and Yang, Xun},
  journal={arXiv preprint arXiv:2508.03034},
  year={2025}
}

@article{wang2024loong,
  title={Loong: Generating minute-level long videos with autoregressive language models},
  author={Wang, Yuqing and Xiong, Tianwei and Zhou, Daquan and Lin, Zhijie and Zhao, Yang and Kang, Bingyi and Feng, Jiashi and Liu, Xihui},
  journal={arXiv preprint arXiv:2410.02757},
  year={2024}
}

@inproceedings{peebles2023scalable,
  title={Scalable diffusion models with transformers},
  author={Peebles, William and Xie, Saining},
  booktitle={Proceedings of the IEEE/CVF international conference on computer vision},
  pages={4195--4205},
  year={2023}
}

@article{guo2025long,
  title={Long context tuning for video generation},
  author={Guo, Yuwei and Yang, Ceyuan and Yang, Ziyan and Ma, Zhibei and Lin, Zhijie and Yang, Zhenheng and Lin, Dahua and Jiang, Lu},
  journal={arXiv preprint arXiv:2503.10589},
  year={2025}
}

@article{cai2025mixture,
  title={Mixture of Contexts for Long Video Generation},
  author={Cai, Shengqu and Yang, Ceyuan and Zhang, Lvmin and Guo, Yuwei and Xiao, Junfei and Yang, Ziyan and Xu, Yinghao and Yang, Zhenheng and Yuille, Alan and Guibas, Leonidas and others},
  journal={arXiv preprint arXiv:2508.21058},
  year={2025}
}

@article{huang2025self,
  title={Self Forcing: Bridging the Train-Test Gap in Autoregressive Video Diffusion},
  author={Huang, Xun and Li, Zhengqi and He, Guande and Zhou, Mingyuan and Shechtman, Eli},
  journal={arXiv preprint arXiv:2506.08009},
  year={2025}
}

@article{teng2025magi,
  title={MAGI-1: Autoregressive Video Generation at Scale},
  author={Teng, Hansi and Jia, Hongyu and Sun, Lei and Li, Lingzhi and Li, Maolin and Tang, Mingqiu and Han, Shuai and Zhang, Tianning and Zhang, WQ and Luo, Weifeng and others},
  journal={arXiv preprint arXiv:2505.13211},
  year={2025}
}

@inproceedings{sun2022dancetrack,
  title={Dancetrack: Multi-object tracking in uniform appearance and diverse motion},
  author={Sun, Peize and Cao, Jinkun and Jiang, Yi and Yuan, Zehuan and Bai, Song and Kitani, Kris and Luo, Ping},
  booktitle={Proceedings of the IEEE/CVF conference on computer vision and pattern recognition},
  pages={20993--21002},
  year={2022}
}

@article{huang2019got,
  title={Got-10k: A large high-diversity benchmark for generic object tracking in the wild},
  author={Huang, Lianghua and Zhao, Xin and Huang, Kaiqi},
  journal={IEEE transactions on pattern analysis and machine intelligence},
  volume={43},
  number={5},
  pages={1562--1577},
  year={2019},
  publisher={IEEE}
}

@inproceedings{xue2022advancing,
  title={Advancing high-resolution video-language representation with large-scale video transcriptions},
  author={Xue, Hongwei and Hang, Tiankai and Zeng, Yanhong and Sun, Yuchong and Liu, Bei and Yang, Huan and Fu, Jianlong and Guo, Baining},
  booktitle={Proceedings of the IEEE/CVF Conference on Computer Vision and Pattern Recognition},
  pages={5036--5045},
  year={2022}
}

@inproceedings{chen2024sharegpt4v,
  title={Sharegpt4v: Improving large multi-modal models with better captions},
  author={Chen, Lin and Li, Jinsong and Dong, Xiaoyi and Zhang, Pan and He, Conghui and Wang, Jiaqi and Zhao, Feng and Lin, Dahua},
  booktitle={European Conference on Computer Vision},
  pages={370--387},
  year={2024},
  organization={Springer}
}

@inproceedings{huang2024vbench,
  title={Vbench: Comprehensive benchmark suite for video generative models},
  author={Huang, Ziqi and He, Yinan and Yu, Jiashuo and Zhang, Fan and Si, Chenyang and Jiang, Yuming and Zhang, Yuanhan and Wu, Tianxing and Jin, Qingyang and Chanpaisit, Nattapol and others},
  booktitle={Proceedings of the IEEE/CVF Conference on Computer Vision and Pattern Recognition},
  pages={21807--21818},
  year={2024}
}

@inproceedings{yin2025causvid,
        title={From Slow Bidirectional to Fast Autoregressive Video Diffusion Models},
        author={Yin, Tianwei and Zhang, Qiang and Zhang, Richard and Freeman, William T and Durand, Fredo and Shechtman, Eli and Huang, Xun},
        journal={CVPR},
        year={2025}
      }

@article{kim2016new,
  title={A new metric of absolute percentage error for intermittent demand forecasts},
  author={Kim, Sungil and Kim, Heeyoung},
  journal={International Journal of Forecasting},
  volume={32},
  number={3},
  pages={669--679},
  year={2016},
  publisher={Elsevier}
}

@article{de2016mean,
  title={Mean absolute percentage error for regression models},
  author={De Myttenaere, Arnaud and Golden, Boris and Le Grand, B{\'e}n{\'e}dicte and Rossi, Fabrice},
  journal={Neurocomputing},
  volume={192},
  pages={38--48},
  year={2016},
  publisher={Elsevier}
}

@inproceedings{li2025longdiff,
  title={LongDiff: Training-Free Long Video Generation in One Go},
  author={Li, Zhuoling and Rahmani, Hossein and Ke, Qiuhong and Liu, Jun},
  booktitle={Proceedings of the Computer Vision and Pattern Recognition Conference},
  pages={17789--17798},
  year={2025}
}

@article{lu2024freelong,
  title={Freelong: Training-free long video generation with spectralblend temporal attention},
  author={Lu, Yu and Liang, Yuanzhi and Zhu, Linchao and Yang, Yi},
  journal={Advances in Neural Information Processing Systems},
  volume={37},
  pages={131434--131455},
  year={2024}
}

@article{han2025show,
  title={Show and polish: reference-guided identity preservation in face video restoration},
  author={Han, Wenkang and Lin, Wang and Zhou, Yiyun and Liu, Qi and Wang, Shulei and Yao, Chang and Chen, Jingyuan},
  journal={arXiv preprint arXiv:2507.10293},
  year={2025}
}

@article{wan2025wan,
  title={Wan: Open and advanced large-scale video generative models},
  author={Wan, Team and Wang, Ang and Ai, Baole and Wen, Bin and Mao, Chaojie and Xie, Chen-Wei and Chen, Di and Yu, Feiwu and Zhao, Haiming and Yang, Jianxiao and others},
  journal={arXiv preprint arXiv:2503.20314},
  year={2025}
}

@misc{dax2025,
  title = {DAX: Diffusion Accelerated eXecution},
  year = {2025},
  publisher = {GitHub},
  journal = {GitHub repository},
  howpublished = {\url{https://github.com/RiseAI-Sys/DAX}}
}

@inproceedings{kwon2023efficient,
  title={Efficient Memory Management for Large Language Model Serving with PagedAttention},
  author={Woosuk Kwon and Zhuohan Li and Siyuan Zhuang and Ying Sheng and Lianmin Zheng and Cody Hao Yu and Joseph E. Gonzalez and Hao Zhang and Ion Stoica},
  booktitle={Proceedings of the ACM SIGOPS 29th Symposium on Operating Systems Principles},
  year={2023}
}

@misc{zheng2024sglangefficientexecutionstructured,
      title={SGLang: Efficient Execution of Structured Language Model Programs}, 
      author={Lianmin Zheng and Liangsheng Yin and Zhiqiang Xie and Chuyue Sun and Jeff Huang and Cody Hao Yu and Shiyi Cao and Christos Kozyrakis and Ion Stoica and Joseph E. Gonzalez and Clark Barrett and Ying Sheng},
      year={2024},
      eprint={2312.07104},
      archivePrefix={arXiv},
      primaryClass={cs.AI},
      url={https://arxiv.org/abs/2312.07104}, 
}

@software{fastvideo2024,
  title        = {FastVideo: A Unified Framework for Accelerated Video Generation},
  author       = {The FastVideo Team},
  url          = {https://github.com/hao-ai-lab/FastVideo},
  month        = apr,
  year         = {2024},
}

@misc{hu2025openrlhfeasytousescalablehighperformance,
      title={OpenRLHF: An Easy-to-use, Scalable and High-performance RLHF Framework}, 
      author={Jian Hu and Xibin Wu and Wei Shen and Jason Klein Liu and Zilin Zhu and Weixun Wang and Songlin Jiang and Haoran Wang and Hao Chen and Bin Chen and Weikai Fang and Xianyu and Yu Cao and Haotian Xu and Yiming Liu},
      year={2025},
      eprint={2405.11143},
      archivePrefix={arXiv},
      primaryClass={cs.AI},
      url={https://arxiv.org/abs/2405.11143}, 
}

@inproceedings{Sheng_2025, series={EuroSys ’25},
   title={HybridFlow: A Flexible and Efficient RLHF Framework},
   url={http://dx.doi.org/10.1145/3689031.3696075},
   DOI={10.1145/3689031.3696075},
   booktitle={Proceedings of the Twentieth European Conference on Computer Systems},
   publisher={ACM},
   author={Sheng, Guangming and Zhang, Chi and Ye, Zilingfeng and Wu, Xibin and Zhang, Wang and Zhang, Ru and Peng, Yanghua and Lin, Haibin and Wu, Chuan},
   year={2025},
   month=mar, pages={1279–1297},
   collection={EuroSys ’25} }

@article{fang2024xdit,
  title={xDiT: an Inference Engine for Diffusion Transformers (DiTs) with Massive Parallelism},
  author={Fang, Jiarui and Pan, Jinzhe and Sun, Xibo and Li, Aoyu and Wang, Jiannan},
  journal={arXiv preprint arXiv:2411.01738},
  year={2024}
}

@misc{yin2024improveddistributionmatchingdistillation,
      title={Improved Distribution Matching Distillation for Fast Image Synthesis}, 
      author={Tianwei Yin and Michaël Gharbi and Taesung Park and Richard Zhang and Eli Shechtman and Fredo Durand and William T. Freeman},
      year={2024},
      eprint={2405.14867},
      archivePrefix={arXiv},
      primaryClass={cs.CV},
      url={https://arxiv.org/abs/2405.14867}, 
}

@misc{zhang2025framecontextpackingdrift,
      title={Frame Context Packing and Drift Prevention in Next-Frame-Prediction Video Diffusion Models}, 
      author={Lvmin Zhang and Shengqu Cai and Muyang Li and Gordon Wetzstein and Maneesh Agrawala},
      year={2025},
      eprint={2504.12626},
      archivePrefix={arXiv},
      primaryClass={cs.CV},
      url={https://arxiv.org/abs/2504.12626}, 
}

@misc{sheng2023flexgenhighthroughputgenerativeinference,
      title={FlexGen: High-Throughput Generative Inference of Large Language Models with a Single GPU}, 
      author={Ying Sheng and Lianmin Zheng and Binhang Yuan and Zhuohan Li and Max Ryabinin and Daniel Y. Fu and Zhiqiang Xie and Beidi Chen and Clark Barrett and Joseph E. Gonzalez and Percy Liang and Christopher Ré and Ion Stoica and Ce Zhang},
      year={2023},
      eprint={2303.06865},
      archivePrefix={arXiv},
      primaryClass={cs.LG},
      url={https://arxiv.org/abs/2303.06865}, 
}

@misc{lee2024infinigenefficientgenerativeinference,
      title={InfiniGen: Efficient Generative Inference of Large Language Models with Dynamic KV Cache Management}, 
      author={Wonbeom Lee and Jungi Lee and Junghwan Seo and Jaewoong Sim},
      year={2024},
      eprint={2406.19707},
      archivePrefix={arXiv},
      primaryClass={cs.LG},
      url={https://arxiv.org/abs/2406.19707}, 
}

@article{liu2024kivi,
  title={KIVI: A Tuning-Free Asymmetric 2bit Quantization for KV Cache},
  author={Liu, Zirui and Yuan, Jiayi and Jin, Hongye and Zhong, Shaochen and Xu, Zhaozhuo and Braverman, Vladimir and Chen, Beidi and Hu, Xia},
  journal={arXiv preprint arXiv:2402.02750},
  year={2024}
}

@misc{li2024snapkvllmknowslooking,
      title={SnapKV: LLM Knows What You are Looking for Before Generation}, 
      author={Yuhong Li and Yingbing Huang and Bowen Yang and Bharat Venkitesh and Acyr Locatelli and Hanchen Ye and Tianle Cai and Patrick Lewis and Deming Chen},
      year={2024},
      eprint={2404.14469},
      archivePrefix={arXiv},
      primaryClass={cs.CL},
      url={https://arxiv.org/abs/2404.14469}, 
}

@misc{zhao2024viditq,
      title={ViDiT-Q: Efficient and Accurate Quantization of Diffusion Transformers for Image and Video Generation}, 
      author={Tianchen Zhao and Tongcheng Fang and Enshu Liu and Wan Rui and Widyadewi Soedarmadji and Shiyao Li and Zinan Lin and Guohao Dai and Shengen Yan and Huazhong Yang and Xuefei Ning and Yu Wang},
      year={2024},
      eprint={2406.02540},
      archivePrefix={arXiv},
      primaryClass={cs.CV}
}

@inproceedings{
  li2024svdquant,
  title={SVDQuant: Absorbing Outliers by Low-Rank Components for 4-Bit Diffusion Models},
  author={Li*, Muyang and Lin*, Yujun and Zhang*, Zhekai and Cai, Tianle and Li, Xiuyu and Guo, Junxian and Xie, Enze and Meng, Chenlin and Zhu, Jun-Yan and Han, Song},
  booktitle={The Thirteenth International Conference on Learning Representations},
  year={2025}
}

@misc{liu2025timestepembeddingtellsits,
      title={Timestep Embedding Tells: It's Time to Cache for Video Diffusion Model}, 
      author={Feng Liu and Shiwei Zhang and Xiaofeng Wang and Yujie Wei and Haonan Qiu and Yuzhong Zhao and Yingya Zhang and Qixiang Ye and Fang Wan},
      year={2025},
      eprint={2411.19108},
      archivePrefix={arXiv},
      primaryClass={cs.CV},
      url={https://arxiv.org/abs/2411.19108}, 
}

@misc{zhang2025fastvideogenerationsliding,
      title={Fast Video Generation with Sliding Tile Attention}, 
      author={Peiyuan Zhang and Yongqi Chen and Runlong Su and Hangliang Ding and Ion Stoica and Zhengzhong Liu and Hao Zhang},
      year={2025},
      eprint={2502.04507},
      archivePrefix={arXiv},
      primaryClass={cs.CV},
      url={https://arxiv.org/abs/2502.04507}, 
}

@misc{fang2024uspunifiedsequenceparallelism,
      title={USP: A Unified Sequence Parallelism Approach for Long Context Generative AI}, 
      author={Jiarui Fang and Shangchun Zhao},
      year={2024},
      eprint={2405.07719},
      archivePrefix={arXiv},
      primaryClass={cs.LG},
      url={https://arxiv.org/abs/2405.07719}, 
}

@inproceedings{fang2024pipefusionpatchlevelpipelineparallelism,
      title={PipeFusion: Patch-level Pipeline Parallelism for Diffusion Transformers Inference}, 
      author={Jiarui Fang and Jinzhe Pan and Jiannan Wang and Aoyu Li and Xibo Sun},
      booktitle={Advances in Neural Information Processing Systems},
      year={2025}
}

@article{liu2024deepseek,
  title={Deepseek-v3 technical report},
  author={Liu, Aixin and Feng, Bei and Xue, Bing and Wang, Bingxuan and Wu, Bochao and Lu, Chengda and Zhao, Chenggang and Deng, Chengqi and Zhang, Chenyu and Ruan, Chong and others},
  journal={arXiv preprint arXiv:2412.19437},
  year={2024}
}

@inproceedings{zhangspargeattention,
  title={SpargeAttention: Accurate and Training-free Sparse Attention Accelerating Any Model Inference},
  author={Zhang, Jintao and Xiang, Chendong and Huang, Haofeng and Xi, Haocheng and Zhu, Jun and Chen, Jianfei and others},
  booktitle={International Conference on Machine Learning},
  year={2025}
}

@inproceedings{yang2025sparse,
  title={Sparse VideoGen2: Accelerate Video Generation with Sparse Attention via Semantic-Aware Permutation},
  author={Yang, Shuo and Xi, Haocheng and Zhao, Yilong and Li, Muyang and Zhang, Jintao and Cai, Han and Lin, Yujun and Li, Xiuyu and Xu, Chenfeng and Peng, Kelly and others},
  booktitle={Advances in Neural Information Processing Systems},
  year={2025}
}

@article{gu2025blade,
  title={BLADE: Block-Sparse Attention Meets Step Distillation for Efficient Video Generation},
  author={Gu, Youping and Li, Xiaolong and Hu, Yuhao and Chen, Minqi and Zhuang, Bohan},
  journal={arXiv preprint arXiv:2508.10774},
  year={2025}
}

@misc{huang2025selfforcingbridgingtraintest,
      title={Self Forcing: Bridging the Train-Test Gap in Autoregressive Video Diffusion}, 
      author={Xun Huang and Zhengqi Li and Guande He and Mingyuan Zhou and Eli Shechtman},
      year={2025},
      eprint={2506.08009},
      archivePrefix={arXiv},
      primaryClass={cs.CV},
      url={https://arxiv.org/abs/2506.08009}, 
}

\clearpage

\end{document}